\documentclass[10pt,twocolumn,letterpaper]{article}

\usepackage{wacv}
\usepackage{times}
\usepackage{epsfig}
\usepackage{graphicx}
\usepackage{amsmath}
\usepackage{amssymb}
\usepackage{booktabs}
\usepackage{multirow}
\usepackage{dblfloatfix}
\usepackage{float}
\usepackage{graphicx}
\usepackage{subcaption}
\usepackage{verbatim}
\def\wacvPaperID{7} % Enter the WACV Paper ID here

\wacvapplicationstrack % Uncomment this line if you are submitting to the Applications Track.

\wacvfinalcopy % *** Uncomment this line for the final submission

\ifwacvfinal
\usepackage[breaklinks=true,bookmarks=false]{hyperref}
\else
\usepackage[pagebackref=true,breaklinks=true,colorlinks,bookmarks=false]{hyperref}
\fi

\begin{document}

%%%%%%%%% TITLE
\title{Detecting Arbitrary Keypoints on Limbs and Skis with Sparse Partly Correct Segmentation Masks}

\author{Katja Ludwig \hspace{1.cm} Daniel Kienzle \hspace{1.cm} Julian Lorenz \hspace{1.cm} Rainer Lienhart\\
Machine Learning and Computer Vision Lab, University of Augsburg\\
{\tt\small \{katja.ludwig, daniel.kienzle, julian.lorenz, rainer.lienhart\}@uni-a.de}
}

\maketitle
\thispagestyle{empty}

%%%%%%%%% ABSTRACT
\begin{abstract}
Analyses based on the body posture are crucial for top-class athletes in many sports disciplines. If at all, coaches label only the most important keypoints, since manual annotations are very costly. This paper proposes a method to detect arbitrary keypoints on the limbs and skis of professional ski jumpers that requires a few, only partly correct segmentation masks during training. Our model is based on the Vision Transformer architecture with a special design for the input tokens to query for the desired keypoints. Since we use segmentation masks only to generate ground truth labels for the freely selectable keypoints, partly correct segmentation masks are sufficient for our training procedure. Hence, there is no need for costly hand-annotated segmentation masks. We analyze different training techniques for freely selected and standard keypoints, including pseudo labels, and show in our experiments that only a few partly correct segmentation masks are sufficient for learning to detect arbitrary keypoints on limbs and skis.
\end{abstract}

%%%%%%%%% BODY TEXT
\section{Introduction}

Video analysis is a commonly used method in many sports disciplines in order to assess the performance and analyze the technique as well as the tactics of the athletes. In individual sports, the position of the body and the sports equipment, if any is used, is at the center of interest. With these measurements, coaches can derive evaluations of actions, body posture, speed, etc., of the athlete. For example, ski jumpers are interested in body and ski angles measured relative to the flight trajectory.

\begin{figure}[t]
  \begin{subfigure}{0.45\linewidth}
\centering    
    \includegraphics[height=4.2cm]{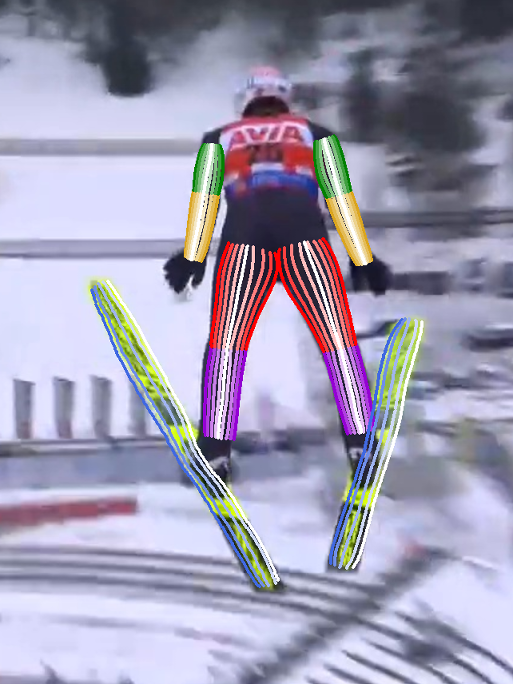}
  \end{subfigure}
  \hfill
  \begin{subfigure}{0.45\linewidth}
\centering    
    \includegraphics[height=4.2cm]{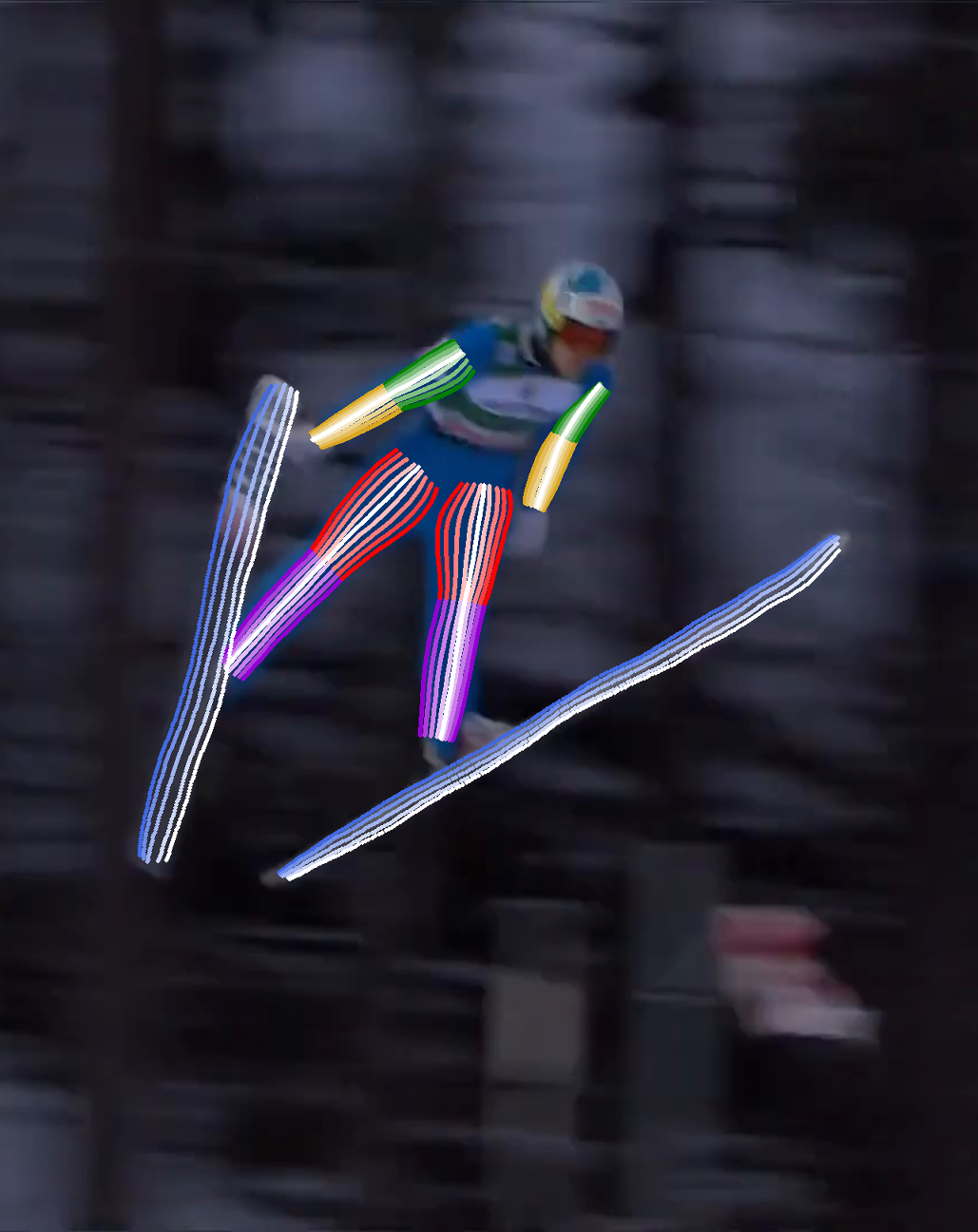}
  \end{subfigure}
  \hfill
   \caption{Two detection results of arbitrary keypoints on the limbs and skis of ski jumpers using our model, visualized with four equally spaced lines to both sides of each limb including the outer boundary in pure color and the central line in white and four equally spaced lines on the skis with a color gradient from one side to the other.}
   \label{fig:example_prediction}
     \vspace{-0.4cm}
\end{figure}

In order to automate the detection of the desired keypoints, 2D Human Pose Estimation (HPE) methods can be used. Their usage reduces the time consumed for video analyses. Therefore, performance analyses are available to more athletes to improve their training outcome. Since annotating a large amount of videos is usually infeasible in the sports domain due to time and money constraints, the datasets for specific sports disciplines are typically very small and comprise only the keypoints that are most important for the analyses. For sports disciplines with human poses that are closely related to standard human activities and keypoints that are common, this problem is solvable with transfer learning. However, images of ski jumpers differ from images of everyday activities, and detecting sports equipment such as skis adds a new complexity. These challenges are reasons why automated keypoint detection is usually bounded to a limited set of standard keypoints. However, the estimation of arbitrary keypoints could open the possibility for more advanced video analyses.

The most common problem of 2D HPE is to estimate the location of a standardized, predefined set of keypoints for each human. A typical size of a keypoint set is around 15 to 30. Often, CNNs are used for the keypoint detection. They involve a backbone network to extract features of the images and a small head network that estimates the location of each predefined standard keypoint. A typical method to predict the keypoint locations is to use a 2D heatmap as the output for each keypoint. Adding additional keypoints would add an additional output channel and require retraining the network head, which makes this approach infeasible for the detection of arbitrary keypoints. Therefore, we use an approach based on a Vision Transformer (ViT) \cite{visiontransformer} architecture. 
%Transformers \cite{transformer} are designed to handle inputs of various length and have recently become popular among vision tasks. 
For 2D HPE, a common ViT architecture is TokenPose \cite{tokenpose}. This method appends a learnable token to the sequence of input image tokens and generates the heatmaps with a small Multi-Layer Perceptron (MLP) from the output of the learnable tokens after the last Transformer layer. Our approach uses the vectorized keypoint approach \cite{ludwig2022recognition} for TokenPose with a slight adaptation for the skis. This approach requires a segmentation mask for every image in order to detect arbitrary points on the limbs of humans. Since many images of ski jumpers are too far from the domain of common segmentation models, we are only capable of collecting a few segmentation masks of body parts that are partly correct and even fewer of the skis. The reason is that we avoid to costly annotate the images by hand but use existing segmentation models \cite{densepose, detectron2} to obtain the masks. We propose and analyze different methods to train a model with this small amount of segmentation masks. Results of our model can be seen in Figure \ref{fig:example_prediction}. The contributions of this work can be summarized as follows:
\begin{itemize}
\item We propose an adapted representation for the vectorized keypoint query tokens introduced by  \cite{ludwig2022recognition} in order to detect arbitrary keypoints on the skis of ski jumpers. 
\item Further, we improve their model such that  keypoint outputs are independent of the number of keypoint tokens in the input sequence.
\item We release a new dataset with 2867 annotated ski jumpers including 13 keypoints on the body and 4 keypoints on the skis together with 424 partly correct segmentation masks sampled from 14 hours of TV broadcast videos during 10 competitions. The dataset is available here: \url{https://www.uni-augsburg.de/en/fakultaet/fai/informatik/prof/mmc/research/datensatze/}
\item We analyze different methods to train a model with only a few partly correct segmentation masks such that it is capable of estimating arbitrary keypoints on limbs and skis of ski jumpers while maintaining similar performance on the standard keypoint set. Our best approach is available here: \url{https://github.com/kaulquappe23/arbitrary-keypoints-skijump} 
\end{itemize}

%-------------------------------------------------------------------------

\section{Related Work}

Analyzing athletes in video footage of training or competition scenarios is common for professional athletes in most sports disciplines. This includes trajectory analysis, e.g. the reconstruction of the 3D trajectory of a badminton shuttle from monocular videos proposed by Liu et al. \cite{badminton}. Furthermore, using the players' trajectories, Wei et al. \cite{wei2015predicting} detect the basketball location from monocular basketball video footage.
In individual sports, the poses of athletes are of great importance. Using the swimming style as an additional input, Einfalt et al. \cite{einfalt2018activity} estimate swimmers' poses and improve the resulting poses by using a pose refinement over time. Furthermore, computer vision is also used in different ski disciplines, mostly for human pose and ski estimation. Wang et al. \cite{wang2019ai} propose a pose correction and exemplar-based visual suggestions to freestyle skiers using human pose estimation. Ludwig et al. \cite{ludwig2020robust} calculate the flight angles for ski jumpers during their flight phase by using  robust estimation methods for human and ski pose recognition. Stepec et al. \cite{skijumpstyle} use estimated poses of ski jumpers and their trajectories in order to automatically generate the style score.

As already mentioned, 2D HPE  is an important method among computer vision based analysis applications in sports. The best scores on common benchmarks like COCO \cite{coco} or MPII Human Pose \cite{mpii} are still based on CNNs \cite{huang2020joint, bulat2020toward}, although Transformer \cite{transformer} based architectures are increasing in popularity. Among CNN architectures, the High Resolution Net (HRNet) \cite{hrnet} is often used, like in \cite{huang2020joint}. It differs from encoder-decoder architectures that are used in former best performing backbones like \cite{maskrcnn, hourglass, simplebaselines} as it keeps large resolution feature maps in the whole network and uses a continuous information exchange between different resolutions. 
Among Transformer \cite{transformer} based HPE approaches, TokenPose \cite{tokenpose} is usable without any convolutions, but using the first stages of an HRNet as feature extractor leads to its best results. TokenPose uses a ViT \cite{visiontransformer}, which embeds small image or feature patches to 1D token vectors serving as the input sequence to the Transformer. Apart from the image patches, learnable keypoint tokens are appended to the input sequence and their output is then transformed through an MLP to heatmaps. Ludwig et al. \cite{ludwig2022detecting} adapt this approach to estimate arbitrary keypoints that lie on the straight line between the fixed keypoints of a dataset and further improve the method to detect freely selected keypoints on the limbs of humans \cite{ludwig2022recognition}. Shi et al. \cite{end2endhpe} propose the first fully end-to-end multi person pose estimation framework based on Transformers. Zeng et al. \cite{tokenclusteringtransformer} cluster the tokens of the Transformer such that less important image areas like the background are represented by less tokens than the humans.

%-------------------------------------------------------------------------
\section{Dataset}\label{sec:dataset}

\begin{figure*}[htb]
  \begin{subfigure}{0.32\linewidth}
\centering    
    \includegraphics[height=3cm]{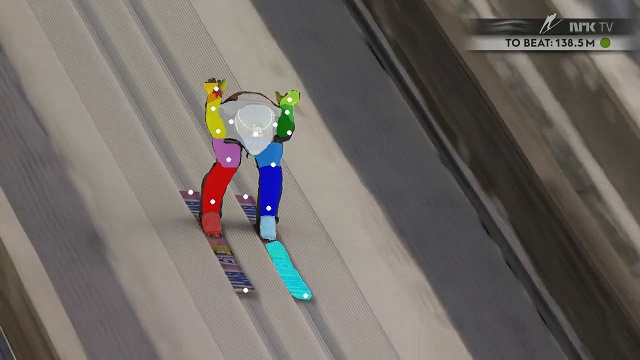}
  \end{subfigure}
  \hfill
  \begin{subfigure}{0.32\linewidth}
  \centering
    \includegraphics[height=3cm]{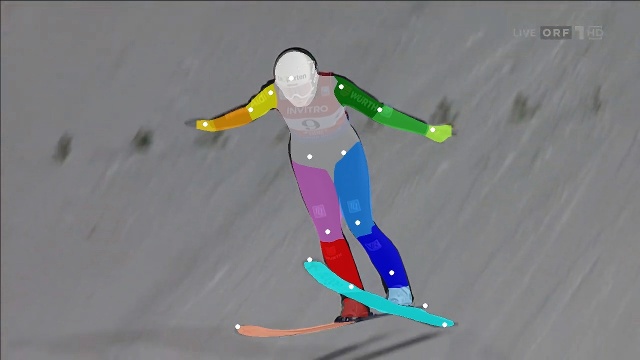}
  \end{subfigure}
  \hfill
  \begin{subfigure}{0.32\linewidth}
  \centering
    \includegraphics[height=3cm]{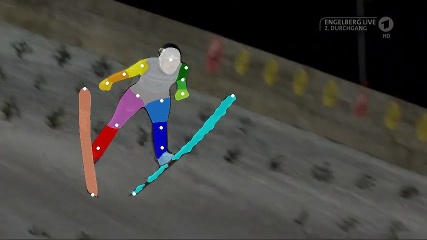}
  \end{subfigure}
  \hfill
  \\
  \begin{subfigure}{0.32\linewidth}
\centering    
    \includegraphics[height=3cm]{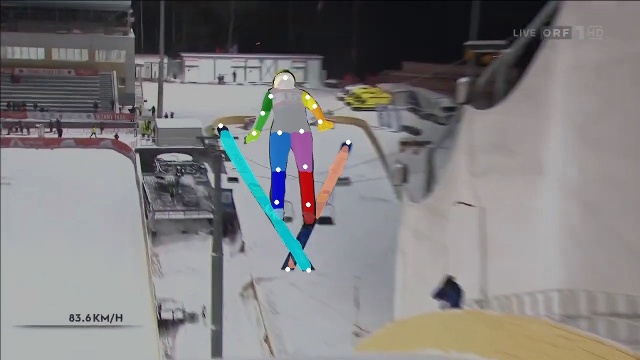}
  \end{subfigure}
  \hfill
  \begin{subfigure}{0.32\linewidth}
  \centering
    \includegraphics[height=3cm]{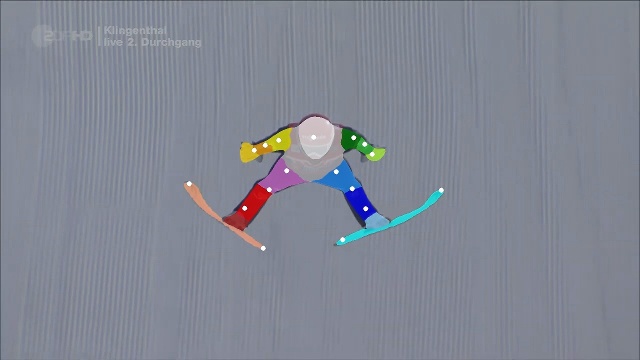}
  \end{subfigure}
  \hfill
  \begin{subfigure}{0.32\linewidth}
  \centering
    \includegraphics[height=3cm]{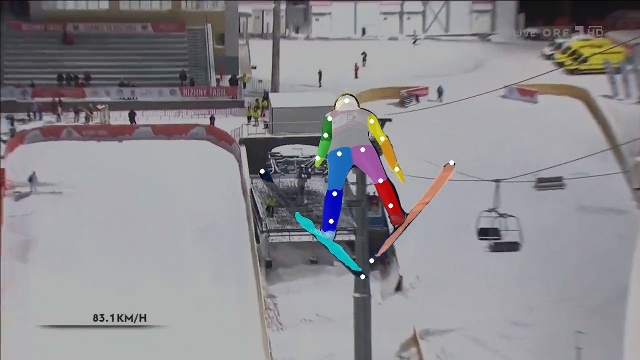}
  \end{subfigure}
  \hfill
  \caption{Example images from our dataset. The images are darkened and the segmentation masks (that are sometimes only partly correct or incomplete) are visualized with an overlay.  Annotated keypoints are displayed with white circles.}
  \label{fig:dataset}
  \vspace{-0.1cm}
\end{figure*}

We collect broadcast TV footage from 10 skijump competitions available on YouTube in order to provide a publicly available dataset for benchmarking arbitrary keypoint detection in ski jumping images. Each video consists of 24 to 62 individual jumps with a total of 370 jumps. We annotate at maximum 8 frames per jump to have a broad diversity of jumps in the dataset. We select images during in-run and during the flight until the moment right before the landing. Over 80\% of the images correspond to the flight phase. The dataset consists of images of various quality and lighting conditions, male and female athletes, and various perspectives of the ski jumpers. We annotate frames during the slowmotion replays as well, since their fidelity is often higher. We include the information if a frame was collected during a slowmotion replay in the dataset. Furthermore, the athlete's names provided in the TV broadcast were collected and added to the dataset. We split the dataset in a train, test and validation subset such that each athlete is only present in one subset. Our dataset consists of 2867 annotations: 2159 for training, 148 for validation, and 560 for testing. The annotated keypoints are head, left/right shoulder, left/right elbow, left/right wrist, left/right hip, left/right knee, left/right ankle, left/right ski tip, left/right ski tail.

We use the detectron2 \cite{detectron2} framework to generate segmentation masks for our dataset. In a first step, we use  DensePose \cite{densepose} to obtain segmentation masks of the body parts. Since images of ski jumpers are far from the domain of DensePose, most of the masks are completely or partly wrong. We select all masks that are mostly correct and discard the other ones, which results in 424 images. As detectron2 is also trained to segment skis, we feed the remaining images through an instance segmentation model in the second step. However, only a small proportion of skis is detected, and even less skis are detected correctly. A second look shows that some skis are detected, but wrongly classified as snowboards, surfboards, etc. Hence, we select and aggregate all masks that belong to skis by hand and split the ski masks in left and right ski. In many cases, only one ski is detected and/or only parts of a ski are contained in the mask. Some example images are displayed in Figure \ref{fig:dataset}. Segmentation masks of the head, torso, left/right upper arm, left/right forearm, left/right hand, left/right thigh, left/right lower leg, left/right foot, and left/right ski are contained in the dataset: 326 segmentation masks in the train subset, 81 in the test subset and only 17 in the validation subset. Because these are too few masks for profound decisions, we coarsely label additional images with the body parts that are of interest for our research (limbs and skis), such that the validation set consists of 46 images.

%-------------------------------------------------------------------------

\section{Method}

\subsection{Architecture Design}

The basis of all models used in this work is the TokenPose-Base architecture \cite{tokenpose}. It is a combined convolutional and Transformer architecture. It uses the first three stages of an HRNet \cite{hrnet} as a first feature extractor. The resulting features of the branch with the highest resolution are then split into feature patches and converted to visual tokens by a linear projection. These tokens are fed to a ViT \cite{visiontransformer}. All methods proposed in this paper are also usable with any other TokenPose variant. 

\subsubsection{Generation of Ground Truth Keypoints}

\begin{figure*}[htb]
  \begin{subfigure}{0.32\linewidth}
\centering    
    \includegraphics[height=3cm]{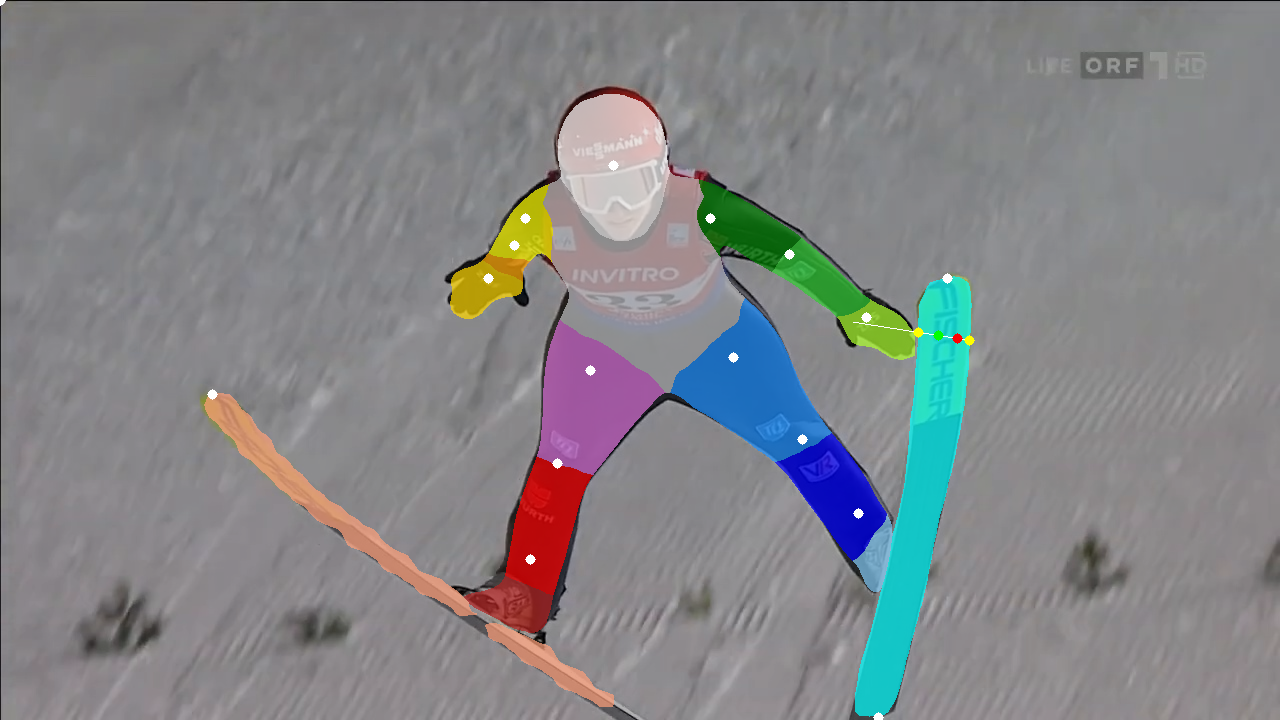}
  \end{subfigure}
  \hfill
  \begin{subfigure}{0.32\linewidth}
  \centering
    \includegraphics[height=3cm]{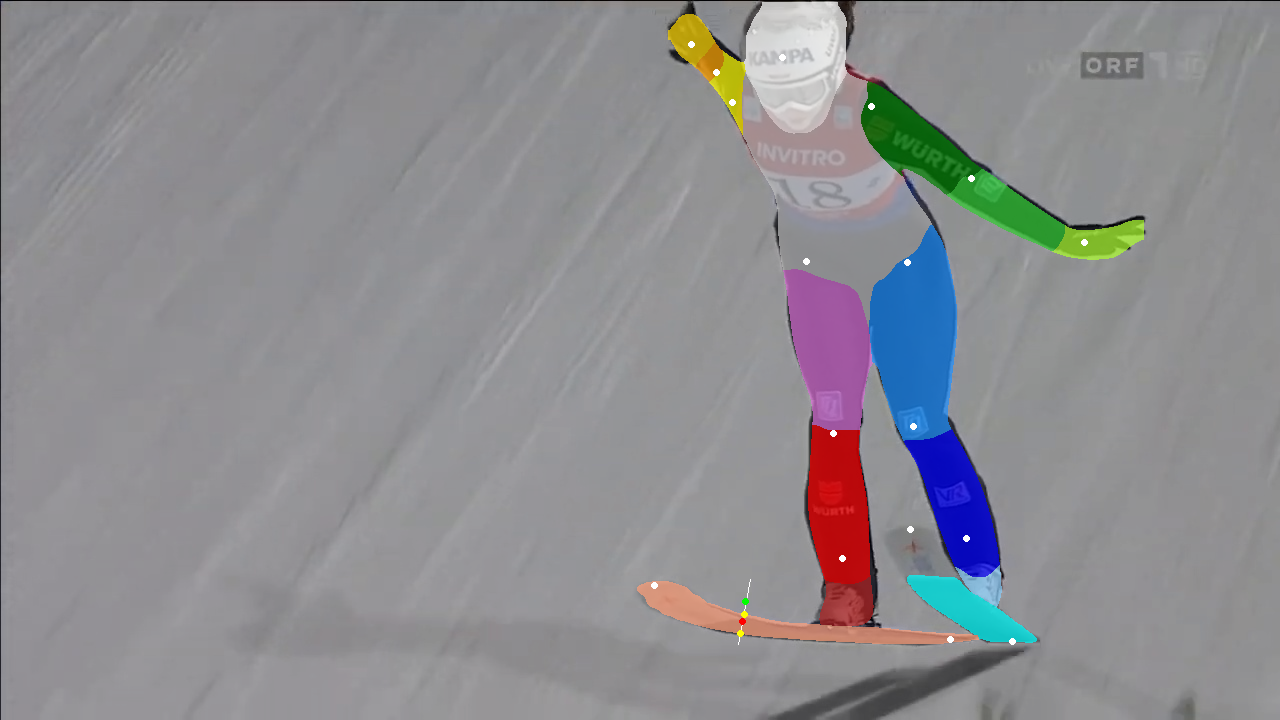}
  \end{subfigure}
  \hfill
  \begin{subfigure}{0.32\linewidth}
  \centering
    \includegraphics[height=3cm]{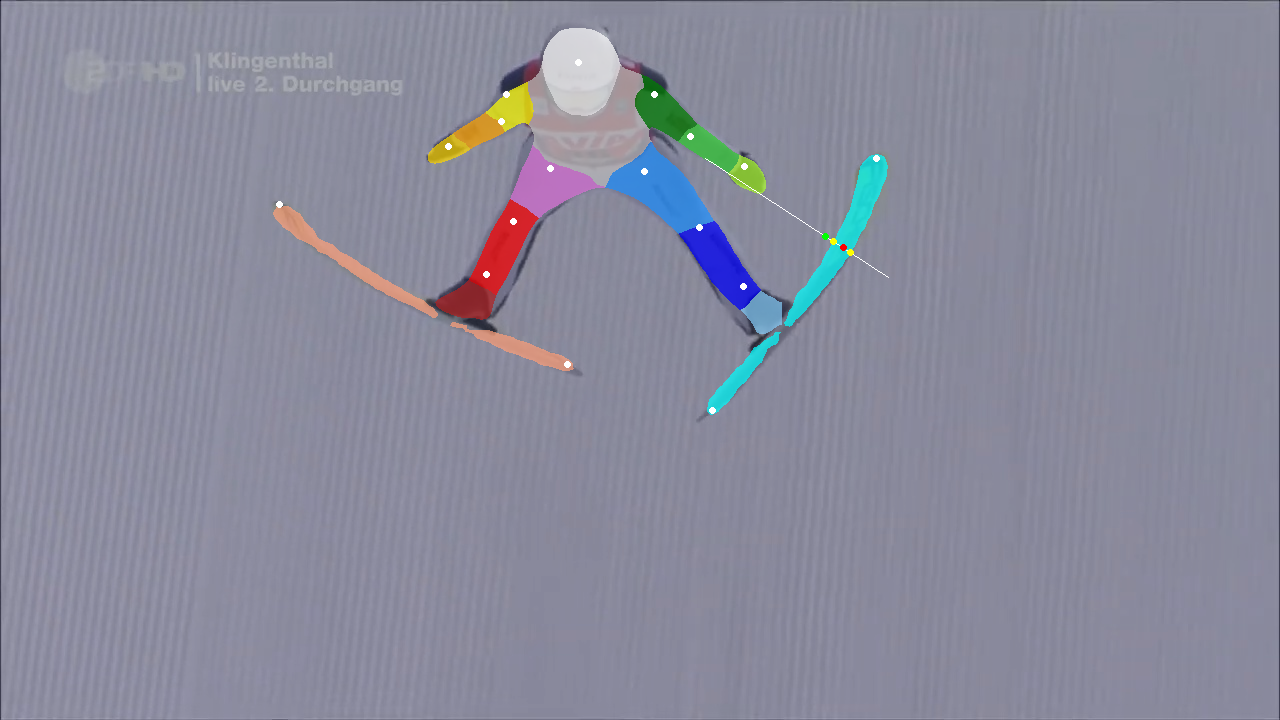}
  \end{subfigure}
  \hfill
  \caption{Examples for the ground truth keypoint generation process on skis. $p$ is visualized in green, the orthogonal line in white, $c_1$ and $c_2$ in yellow and the generated point in red. The middle and right image show that the projection point can be located outside of a ski mask.}
  \label{fig:kp_gen}
  \vspace{-0.3cm}
\end{figure*}

We follow the strategy presented in \cite{ludwig2022recognition} to generate labels for arbitrary keypoints on the limbs. At first, a random point is selected on the straight line between two keypoints enclosing a body part, called the projection point $p$. Second, a line orthogonal to this straight line through $p$ is created. This line has two intersection points $c_1$ and $c_2$ with the border of the segmentation mask of the corresponding body part. Now, a random point $r$ either between $c_1$ and $p$ or between $c_2$ and $p$ is selected such that both sides are equally likely and more points lie close to the intersection points, as these keypoints are more difficult for the model to detect.

However, this method is not directly applicable for arbitrary points on skis because the straight lines between ski tips and ski tails do not necessarily lie entirely within the segmentation mask of the skis, depending on the perspective and the bending of the skis. As a consequence, $p$ might also lie outside the segmentation mask of the ski. See the middle and right images of Figure \ref{fig:kp_gen} for examples. Hence, we randomly select a point $r$ between $c_1$ and $c_2$, with a high probability that the point is located close to one of the intersection points and a lower probability that it lies in the middle, as evaluations show that it is harder for the model to learn the keypoints close to the boundary.

\subsubsection{Keypoint Query Tokens}\label{sec:keypoint_token}

TokenPose is only able to detect the fixed keypoints defined for each dataset. For each keypoint, it learns a token that is appended to the sequence of visual tokens created by extracting features from the input image or the input image directly. As shown in \cite{ludwig2022detecting}, these tokens exhibit no correlation. Therefore, it is necessary to redesign the tokens to control their meaning. In order to represent arbitrary tokens on the segmentation masks, we use the \emph{vectorized keypoints} approach presented in \cite{ludwig2022recognition} for the limbs and adapt it for skis. 
Hence, a keypoint vector $v^p$ and a thickness vector $v^t$ are designed and converted via a learned linear projection to a vector of half of the embedding size used in the ViT. Then, $v^p$ and $v^t$ are concatenated to a single keypoint query token. All keypoint query tokens are appended to the sequence of visual tokens and then fed jointly through the Transformer network. A positional encoding is added only to the visual tokens after each Transformer layer. After the last layer, a small MLP with shared weights is used to convert the keypoint query tokens to heatmaps. 

For a dataset with $n$ keypoints, the projection point is encoded as a vector $v^p \in \mathbb{R}^n$. Let $k_i$, $k_j$ be the keypoints enclosing the body part. Then, each projection point $p$ can be formalized as $p = \alpha k_i + (1-\alpha)k_j$ and $v^p$ is created as
\begin{equation}
v^p_h =\left\{\begin{array}{ll} 1-\alpha, & h = j\\
										\alpha, & h = i\\
         								0, & h \neq i \land h \neq j\end{array}\right .  h = 1, ..., n
\end{equation}
If a standard keypoint should be detected, $\alpha = 1$. The position of an arbitrary keypoint $r$ is now encoded relative to $p$ which we call thickness. If $r$ is a point on a limb, it can be formalized as $r = \beta p + (1-\beta) c_{1/2}$, with $c_{1/2}$ being the intersection point closer to $r$. If $r$ is a point on the skis, it can be formalized as $r = \beta c_1 + (1-\beta) c_2$. Furthermore, we define the thickness vector $v^t \in \mathbb{R}^3$ as
\begin{equation}
v^t =\left\{\begin{array}{ll} \left(1-\beta, \beta, 0 \right)^T  , & r \text{ is on limb} \land c_{1/2} = c_1\\
										\left(0, \beta, 1-\beta \right)^T  , & r \text{ is on limb} \land c_{1/2} = c_2\\
										\left( \beta,  0, 1- \beta\right)^T  , & r  \text{ is on ski} \end{array}\right.  
\end{equation}
For standard points on limbs \textbf{and} skis, $(0, 1, 0)^T$ is used.

\subsubsection{Attention Targets}\label{sec:attention}

\begin{figure*}[htb]
  \begin{subfigure}{0.19\linewidth}
	\centering    
    \includegraphics[height=3.4cm]{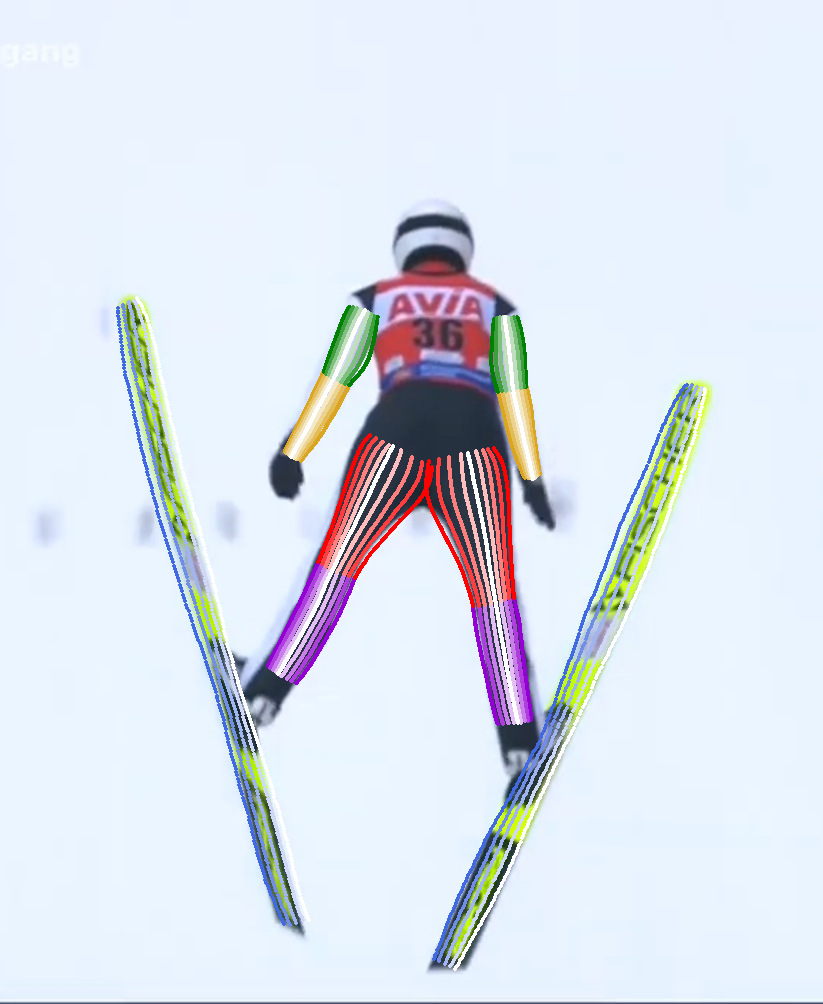}
     \caption{Adapted att. (ours)}
  \end{subfigure}
  \hfill
  \begin{subfigure}{0.19\linewidth}
  \centering
    \includegraphics[height=3.4cm]{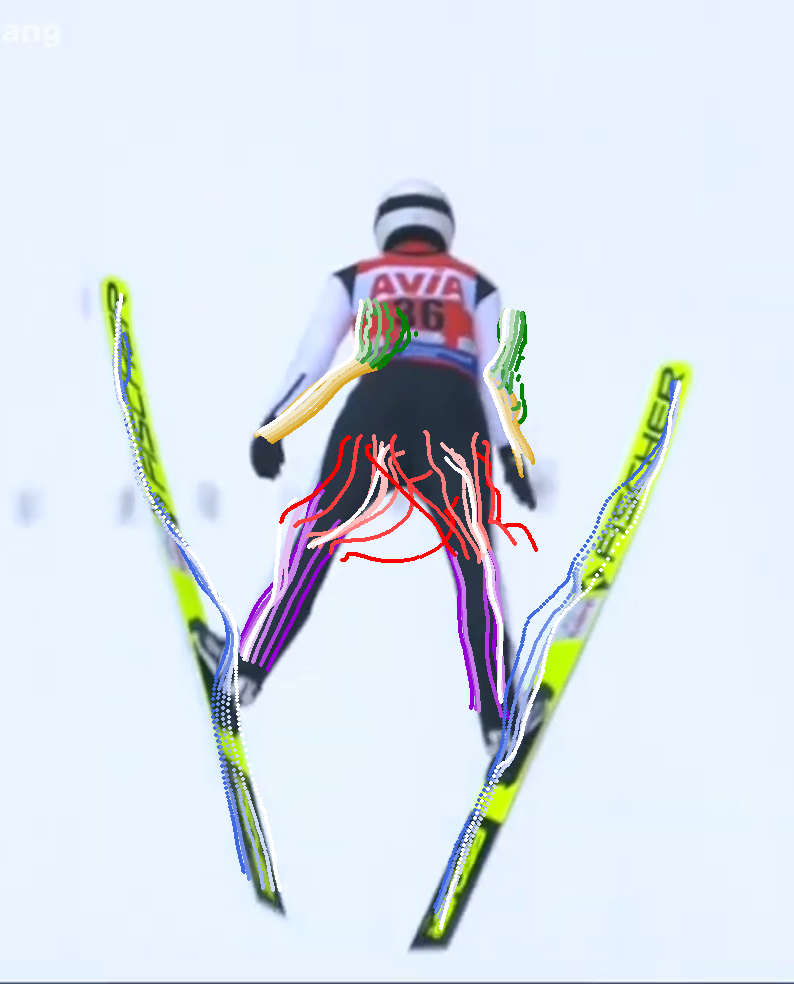}
     \caption{All points}
  \end{subfigure}
  \hfill
  \begin{subfigure}{0.19\linewidth}
  \centering
    \includegraphics[height=3.4cm]{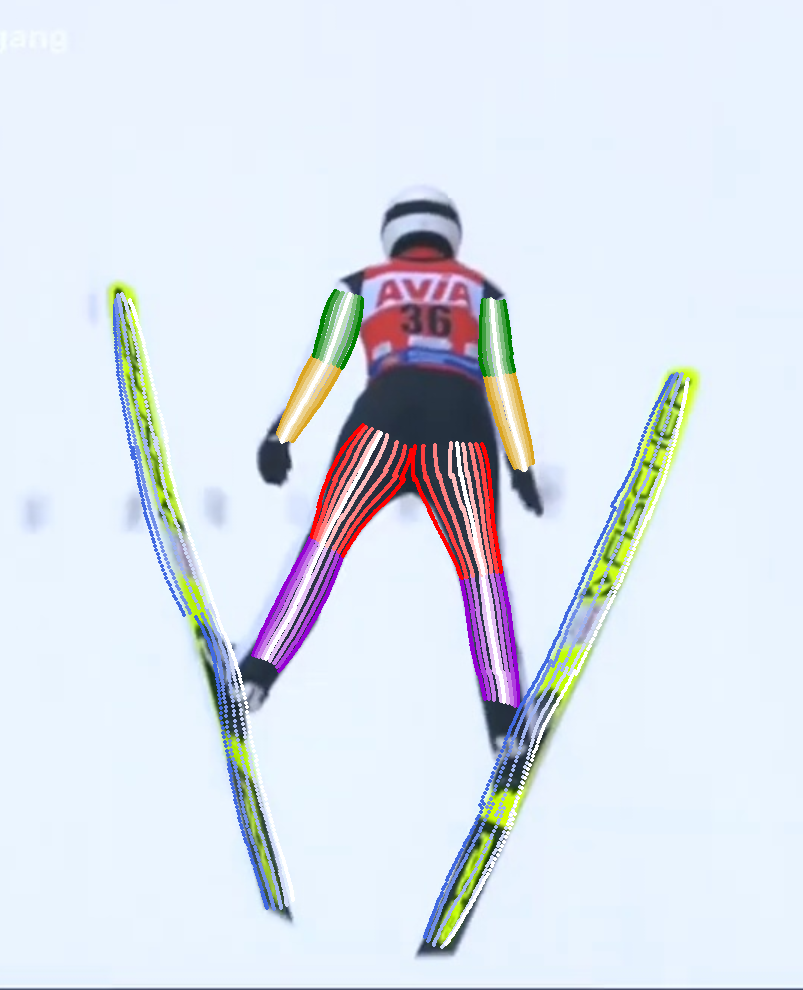}
     \caption{50 points}
  \end{subfigure}
  \hfill
  \begin{subfigure}{0.19\linewidth}
  \centering
    \includegraphics[height=3.4cm]{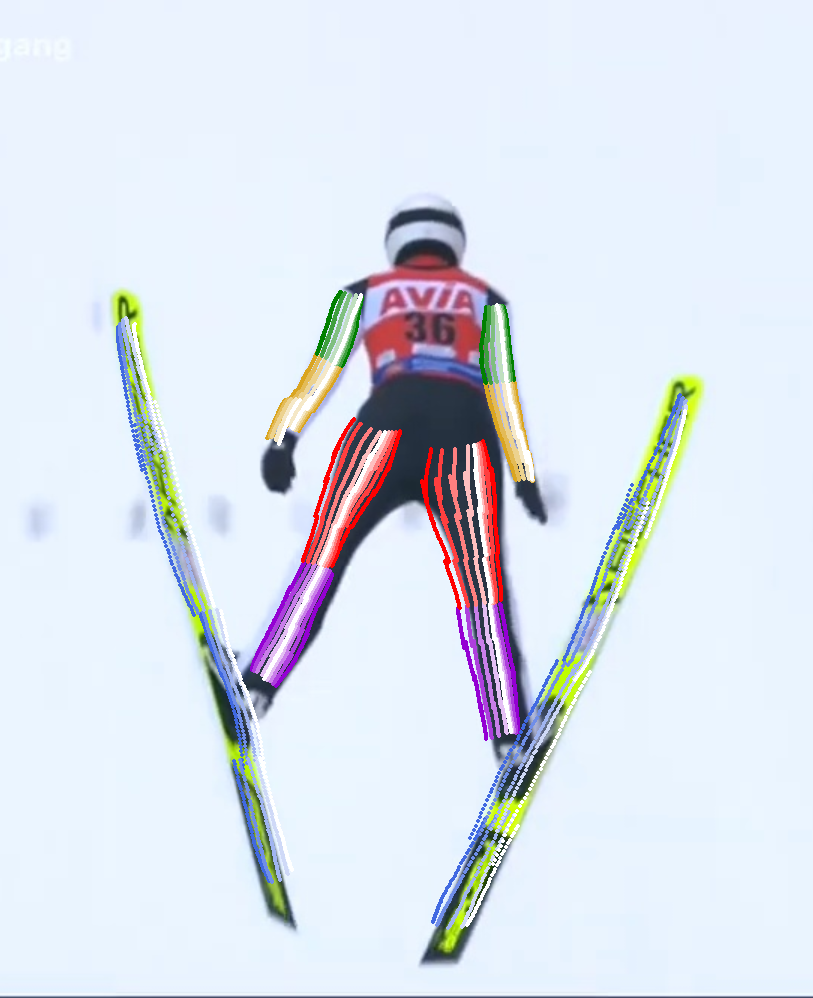}
    \caption{10 points}
  \end{subfigure}
  \hfill
  \begin{subfigure}{0.19\linewidth}
  \centering
    \includegraphics[height=3.4cm]{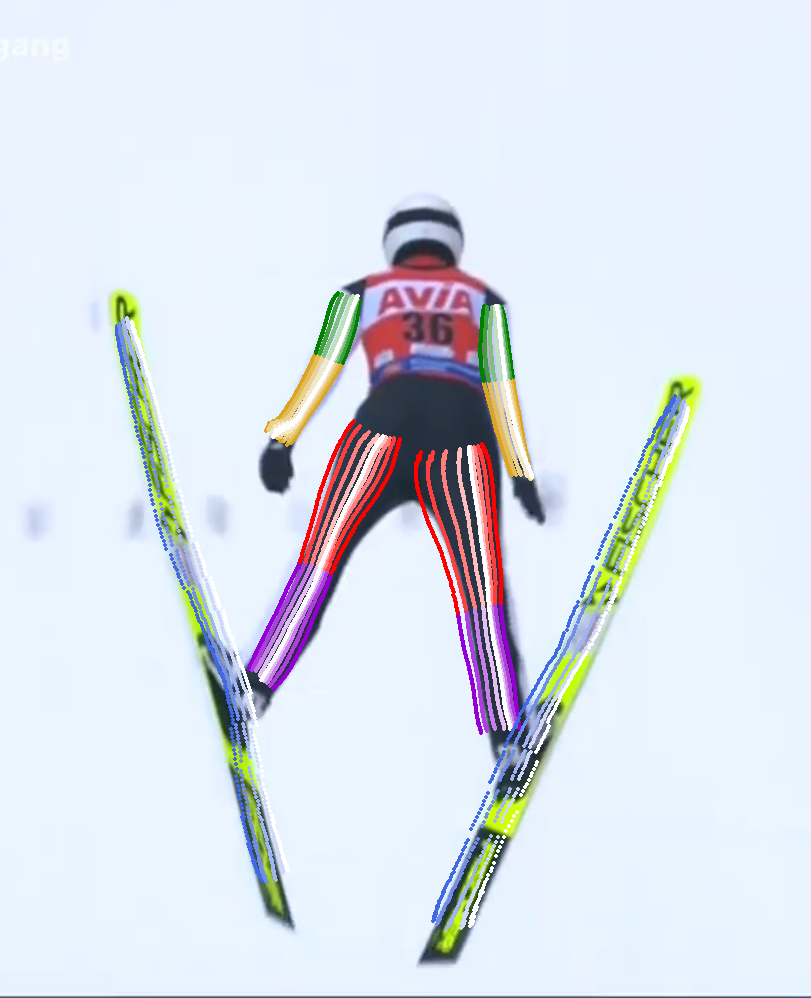}
    \caption{50 points w/o masking}
  \end{subfigure}
  \hfill
    \vspace{-0.2cm} 
  \caption{Examples for model detections depending on the number of keypoint query tokens per model call. The images show four equally spaced lines regarding the thickness on each body part. For the limbs, the projection line is colored white with a color gradient to the edges. For the skis, the color gradient is from one side to the other. The keypoint query tokens are identical for all images. Image (a) is the result for the adapted attention, independent of the number of keypoint queries per model execution and without random sampling. Images (b) - (d) use the original attention with random sampling like \cite{ludwig2022recognition}, image (e) without random sampling. In image (b), all keypoints are computed with one model execution. In image (c) and (e), only 50 points are computed in one inference step and in image (d), 10 points are computed at once.}
  \label{fig:attention_examples}
  \vspace{-0.3cm}
\end{figure*}
Evaluations show that the method presented in \cite{ludwig2022recognition} works well if the number of keypoint query tokens used during inference is similar to the number of keypoint query tokens used during training. If a lot more tokens are used, the detection performance decreases. See Figure \ref{fig:attention_examples} for some examples. Thus, the model output for one keypoint query affects the other queries, which is an undesired behavior. The reason for this behavior is the attention mechanism. In TokenPose, the attention of layer $i+1$ is calculated as
\begin{equation}
A(L^{i+1}) = softmax(\frac{L^iW_Q(L^iW_K)^T}{\sqrt{d}})(L^iW_V)
\end{equation}
where $L^i = (T^i_{vis}, T^i_{kp})$ are the visual and keypoint query tokens of the previous layer, $W_Q, W_K, W_V \in \mathbb{R}^{d\times d}$ are the learnable parameters to generate the queries, keys and values and $d$ is the dimension of the tokens. Hence, the attention is calculated between all tokens, so there is an information flow from the keypoint query tokens to the visual tokens. Therefore, the keypoint query tokens have an influence on each other directly and through the visual tokens. In TokenPose, this is a desired behavior, as always the same keypoints are detected and the information of other keypoint tokens can help to detect occluded keypoints \cite{tokenpose}. In \cite{ludwig2022detecting}, it is observed that the detection performance is decreasing if the keypoint tokens corresponding to the standard keypoints are always present during training, but left out during inference.  Their solution is to include a random sampling and permutation of the keypoint query tokens, but this does not solve the problem of the undesired influence of the keypoint tokens on each other. 
\begin{figure}[b]
  \vspace{-0.4cm}  
  \centering
  \begin{subfigure}{\linewidth}
\centering    
    \includegraphics[width=\linewidth]{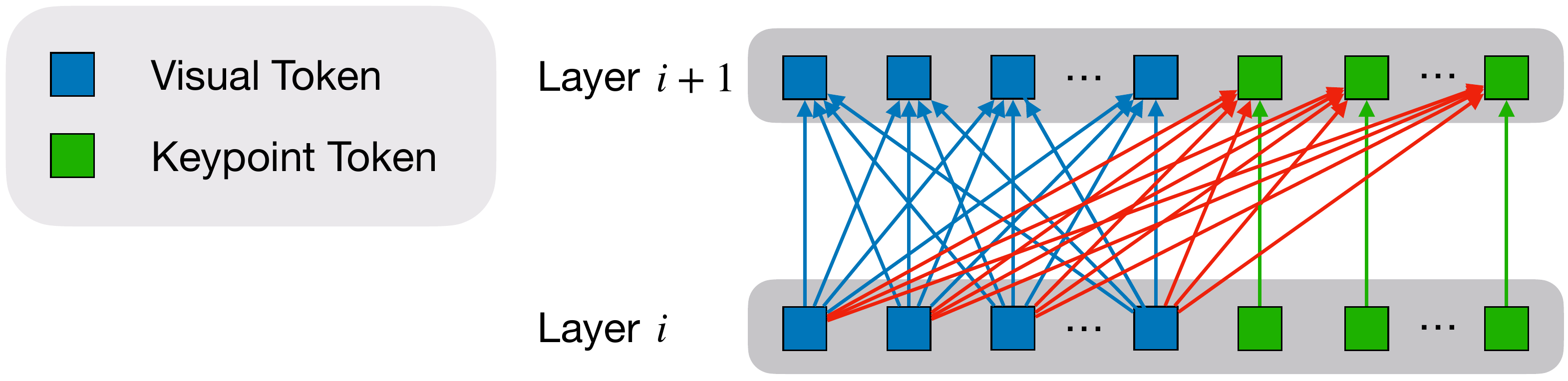}
  \end{subfigure}
  \hfill
  \vspace{-0.3cm}
   \caption{Visualization of the information flow in our adapted attention modules. The attention is computed such that only the visual tokens serve as keys and values. Hence, the visual tokens exchange information and keypoint tokens aggregate information from the visual tokens. 
   The keypoint tokens do not influence other tokens.
   }
   \label{fig:attention}
\end{figure}
Therefore, we adopt the attention mechanism according to
\begin{equation}
\widehat{A}(L^{i+1}) = softmax(\frac{L^iW_Q(T^i_{vis}W_K)^T}{\sqrt{d}})(T^i_{vis}W_V)
\end{equation}
 which is also visualized in Figure \ref{fig:attention}. The keypoint tokens serve only as the queries during the attention and the visual tokens as queries, keys and values. This strategy is similar to \cite{crossattention}. The information flow in the Transformer network is now restricted within the visual tokens and from visual tokens to keypoint tokens. Hence, the position of a detected keypoint is only dependent on the image and independent of the other keypoints that should be detected at the same time. This is the desired behavior.
Furthermore, the dimension of the softmax is now fixed to the number of visual tokens and independent of the number of keypoint tokens.

\subsection{Training Strategies}

As a baseline, we train a model with our adaptations on the images with available segmentation masks. As we only have a few masks available and the train subset is thus small, this model underperforms regarding standard keypoints. Therefore, we evaluate different strategies for including the full dataset in the arbitrary keypoint training. One approach could be the finetuning of a segmentation net with our segmentation masks in order to generate segmentation masks for all images. However, this approach is infeasible as the segmentation masks of our dataset are only partially complete and only partially correct. Finetuning a segmentation net on these masks would not let the model learn useful masks. This is especially the case for skis because many images have annotations for only one, no ski, or only parts of the skis. For a direct training on arbitrary points, this is not a problem. Arbitrary points are only generated on available (possibly partial) segmentation masks. This does not deteriorate the model's performance. The only challenge are segmentation masks with incorrect borders, since this leads either to a wrong calculation of the intersection points and a mismatch of the thickness vector and the generated point or to a generated point that does not lie on the limb/ski. However, our experiments show that the model can cope with this challenge and learns the correct points in most cases, because the number of correctly created points is by far larger than the number of false points.

\subsubsection{Combined Training of Arbitrary and Standard Keypoints}\label{sec:train_strategy}

The most straightforward approach is to use all available images for the training on the standard keypoints and the segmentation masks for the arbitrary keypoints. This strategy increases the performance on the standard keypoints a lot, but also deteriorates the ability to detect arbitrary points.
Another technique includes the detection of projection points as presented in \cite{ludwig2022detecting}. In order to generate arbitrary keypoints on the straight line between two keypoints, which we call projection points, segmentation masks are not necessary. Hence, we can use all images for training on standard and projection points and jointly train with arbitrary points on the images with available segmentation masks. 

\subsubsection{Pseudo Labels}\label{sec:pl}

The aforementioned approaches include the full train set during training, but the model still learns to detect arbitrary points only from a small subset. Therefore, we also experiment with pseudo labels. This means that we use a trained model in order to generate labels of arbitrary points for all images. With this strategy, it is possible to train a model on arbitrary keypoints with the whole training set. After convergence on the pseudo labels, a finetuning is executed with arbitrary points generated from the available segmentation masks, because these ground truth keypoints are more precise than the pseudo labels. Another strategy is to add the pseudo label training as a third part to the already described combined training approaches. 

Looking at visualizations of the generated pseudo labels reveals some wrong pseudo labels. Furthermore, we observe that the network's scores have no direct relation to the correctness of the pseudo labels. Hence, we use another technique to filter the labels. First, we obtain the model's predictions from the original image and some augmented variants. Second, we remove all keypoints with low scores, since a low score should indicate that a keypoint is not visible and augmentations like rotations might move the keypoint outside of the augmented image. We remove all keypoints from the pseudo labels with too few detections. Next, we transform the detections belonging to the augmented versions back to their location in the original image. Then, we calculate the standard deviation of these keypoints relative to the torso size. We use the standard deviation as the confidence measure instead of the network score. We select the pseudo labels with the least standard deviation per body part, that the number of pseudo labels per body part is equal in the pseudo label dataset. This approach is similar to \cite{ludwig2021self}.

%-------------------------------------------------------------------------

\section{Experiments}

The backbone for all experiments is TokenPose-Base \cite{tokenpose} with three stages of an HRNet-w32 \cite{hrnet} for feature extraction. We crop the ski jumpers and resize all images to $256 \times 192$. Cropping is performed by creating the tightest bounding box containing all standard keypoints and enlarging its width and height by 20\% to all sides. Visual and keypoint tokens are of size $192$. We use 2D sine as positional encoding. After the Transformer layers, we use a two-layer MLP to obtain heatmaps of size $64 \times 48$ from the keypoint tokens. Keypoints coordinates are obtained from the heatmaps via the DARK \cite{dark} method. We pretrain our models with the COCO \cite{coco} dataset, either with TokenPose - only on the standard keypoints -  or with the vectorized keypoints approach using arbitrary keypoints on the limbs. Additional to the model that is being trained, we keep an Exponential Moving Average (EMA) of the model's weights with an EMA rate of 0.99. The EMA model behaves like a temporal ensemble and achieves slightly better results than the original model. Therefore, we evaluate all experiments with the EMA model. As described in Section \ref{sec:dataset}, we evaluate on the test set with 560 images in total and 81 images with segmentation masks. We generate 200 arbitrary keypoints with a fixed seed for each image during the arbitrary keypoint evaluation, resulting in 16,200 total keypoints.

\subsection{Evaluation Metrics}

The first evaluation metric that we use is the Percentage of Correct Keypoints (PCK). A keypoint is considered as correct according to the PCK at a certain threshold $t$, if the euclidean distance between the ground truth keypoint and the detected keypoint is less or equal than $t$ times the torso size. We use the euclidean distance between right shoulder and left hip joint as the torso size and a threshold of 0.1. For this dataset, this threshold corresponds to approx. 6cm.

\begin{table*}[htb]
\begin{center}
\resizebox{0.99\linewidth}{!}{ 
\begin{tabular}{c|c|cccc|cccc}
\toprule
Method & Pretraining & Std. KP & Seg. M. & Proj. KP & PL & Std. PCK  & Full PCK & MTE $\downarrow$ & PCT $\uparrow$\\
\midrule
 TokenPose & Std. KP & \checkmark & & & & 77.2\\ 
  TokenPose & VK & \checkmark & & & & 75.4\\ 
 \midrule
  Vectorized Keypoints & VK & & \checkmark & & & 52.7 & 88.1 & 18.2 & \textbf{77.7}\\
  \midrule
  Std. \& Seg. & VK & \checkmark & \checkmark & & & 77.1 & 90.1 & 18.3 & 76.6\\
Seg. \& Proj. & VK &  & \checkmark & \checkmark & & 76.5 & \textbf{91.8} & \textbf{17.5} & \textbf{77.7} \\
Seg. \& Proj. & Std. KP &  & \checkmark & \checkmark & & \textbf{77.8} & 91.5 & 18.0 & 76.4 \\
all PL & VK & & & & all &76.3 & 90.4 & 18.7 & 76.0 \\
finetune all PL & VK & & \checkmark & & all & 76.3 & 90.4 & 18.9 & 75.2\\
all PL \& Std.\& Seg. & VK & \checkmark & \checkmark & & all & 76.4 & 90.9 & 19.2 & 74.7 \\
all PL \& Proj. \& Seg. & VK &  & \checkmark & \checkmark  & all &  76.9 & 91.0 & 18.4 & 75.6\\
80\% PL & VK & & & & 80\% & 76.3 & 90.8 & 18.7 & 74.8\\
finetune 80\% PL & VK & & \checkmark & & 80\% & 76.1 & 91.3 & 18.3 & 75.8 \\
80\% PL \& Std.\& Seg. & VK & \checkmark & \checkmark & & 80\% & 77.3& 90.7 & 19.4 & 73.4\\
80\% PL \& Proj.\& Seg. & VK &  & \checkmark & \checkmark  & 80\% & 76.7& 91.4 & 18.1 & 75.3 \\
 \bottomrule
\end{tabular}
}
\vspace{-0.5cm}
\end{center}
\caption{Recall values for the skijump test set in \% at PCK@$0.1$. 
The second column displays the pretraining, \emph{Std. KP} refers to the pretraining with the standard keypoints, \emph{VK} to the pretraining with the vectorized keypoints approach, both on the COCO dataset.
The third table section shows the used training steps. \emph{Std. KP} means training on the standard keypoints, usable on the whole training set. \emph{Seg. M.} refers to the training on arbitrary keypoints with available segmentation masks. \emph{Proj. KP} stands for the training on the projection keypoints which is also usable on the whole training set and \emph{PL} refers to the pseudo labels, whereby either all pseudo labels are used or the 80\% with the least standard deviation during filtering.
The first column of the last table section displays the average PCK of the standard keypoints, evaluated on the test set containing images with and without segmentation masks. The average PCK score including the arbitrary points is given in the second column, the third column shows the MTE and the last column the PCT at threshold $0.2$. These scores are evaluated on the test set with available segmentation masks. } \label{tab:results}
\vspace{-0.4cm}
\end{table*}

We use the terminology \emph{thickness} for the distance between a keypoint and its projection point. As described in \cite{ludwig2022recognition}, the PCK is not sufficient to measure if the model predicts the thickness of the arbitrary points correctly. A model predicting only the projection points would achieve a high PCK score although the thickness might be wrong, because the projection points are close enough to the ground truth points. Therefore, like in \cite{ludwig2022recognition}, the Mean Thickness Error (MTE) and the Percentage of Correct Thickness (PCT) are used as additional metrics. 
%For keypoints on the limbs, the desired thickness is calculated as the ratio of the ground truth keypoint's thickness to the the thickness of the corresponding intersection point $c_{1/2}$. If the detected keypoint is located on the correct side of the projection point, the estimated thickness is calculated analogous. If the keypoint is located on the wrong side, the estimated thickness is calculated as the desired thickness plus the ratio of the estimated keypoint's thickness to the thickness of the wrong intersection point. The thickness error is now the absolute difference between desired and estimated thickness. 
Let $g$ be the ground truth keypoint, $d$ the detected keypoint, $p$ the projection point, $c_g$ the intersection point closer to $g$ and $c_d$ the intersection point closer to $d$. Then, for keypoints on the limbs, the desired thickness $t_{g}$ and the estimated thickness $t_d$ are calculated as
\begin{equation}
t_{g} = \frac{||p - g||_2}{||p - c_g||_2},\;t_d =\left\{\begin{array}{ll}  \frac{||p - d||_2}{||p - c_g||_2}, &c_g = c_d \\[8pt]
         																							\frac{||p - d||_2}{||p - c_d||_2} + t_g, &c_g \neq c_d\end{array}\right .
\end{equation}
The thickness error $e$ is defined as $e = |t_g - t_d|$. Hence, the maximum thickness error is 2, which is set for estimated keypoints that are located outside of the corresponding segmentation mask. In this case, projection and intersection points can not be computed. For arbitrary keypoints on the skis, we adapt this metric to fit the slightly different thickness logic as described in Section \ref{sec:keypoint_token}. Let $g$ be the ground truth keypoint, $d$ the detected keypoint, $c_1$ and $c_2$ the intersection points, then the desired thickness $t_{g}$ and the estimated thickness $t_d$ are calculated as
\begin{equation}
t_{g} = \frac{||c_1 - g||_2}{||c_1 - c_2||_2},\quad t_{d} = \frac{||c_1 - d||_2}{||c_1 - c_2||_2}
\end{equation}
The MTE metric is the mean of all thickness errors and the PCT is defined analogous to the PCK. At threshold $t$, the PCT considers all estimated thicknesses as correct with a thickness error less or equal than $t$. As the maximum thickness error is 2, we use the PCT@0.2 for our evaluations.

\subsection{Results}\label{sec:results}

\newcommand{\mysize}{3.2cm}
\begin{figure*}[htb]
\begin{subfigure}{0.15\linewidth}
	\centering    
    \includegraphics[height=\mysize]{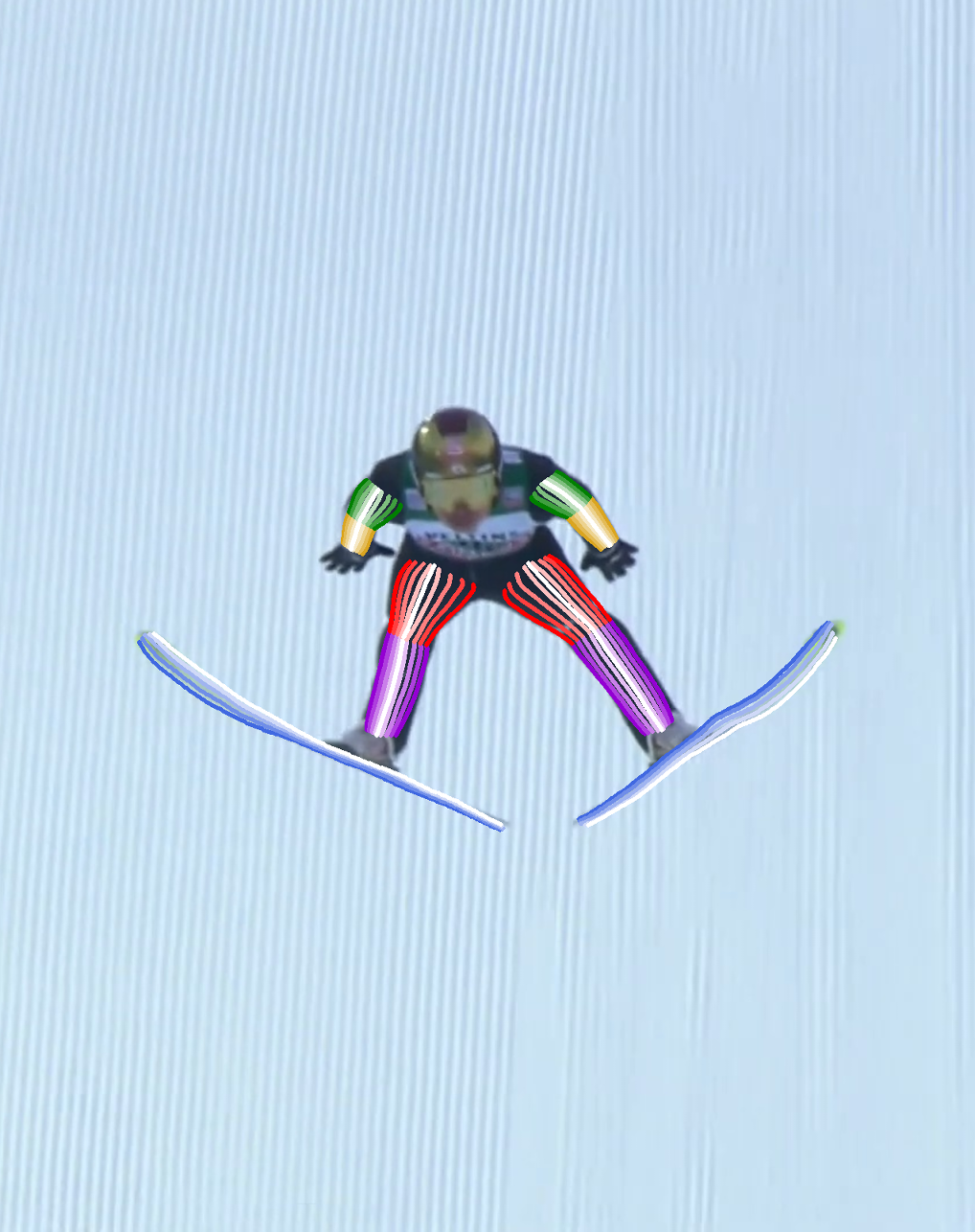}
  \end{subfigure}
  \hfill  
  \begin{subfigure}{0.15\linewidth}
	\centering    
    \includegraphics[height=\mysize]{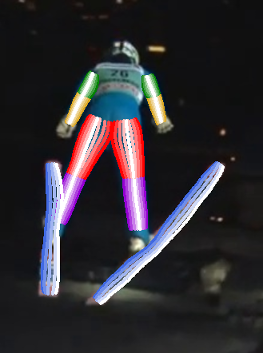}
  \end{subfigure}
  \hfill
  \begin{subfigure}{0.15\linewidth}
  \centering
    \includegraphics[height=\mysize]{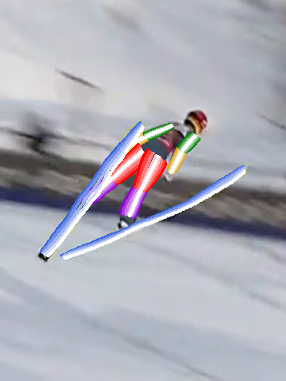}
  \end{subfigure}
  \hfill
  \begin{subfigure}{0.15\linewidth}
  \centering
    \includegraphics[height=\mysize]{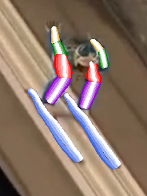}
  \end{subfigure}
  \hfill
  \begin{subfigure}{0.15\linewidth}
  \centering
    \includegraphics[height=\mysize]{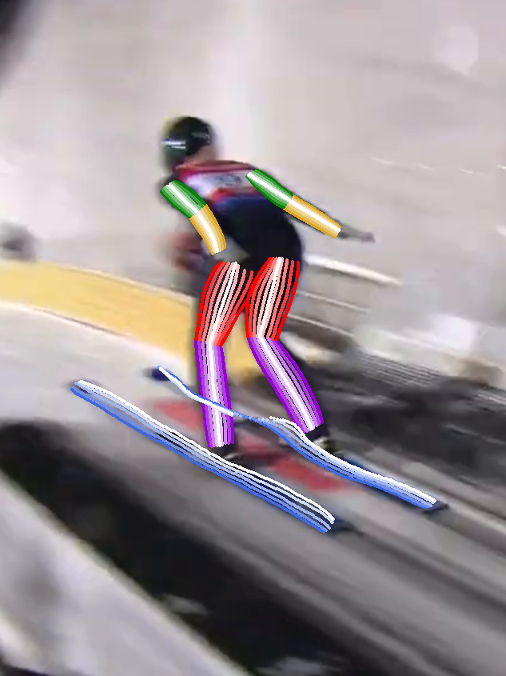}
  \end{subfigure}
  \hfill
  \begin{subfigure}{0.18\linewidth}
  \centering
    \includegraphics[height=\mysize]{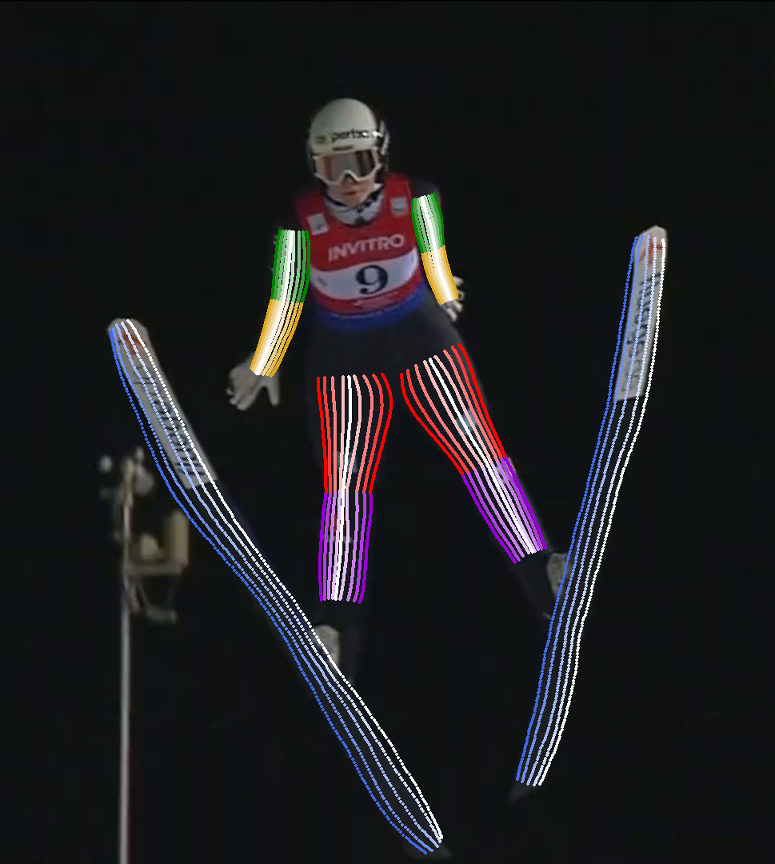}
  \end{subfigure}
  \hfill
  \vspace{-0.2cm}
  \caption{Qualitative examples for model detections. The images show four equally spaced lines regarding the thickness on each body part. For the limbs, the projection line is colored white with a color gradient from it to the edges. For the skis, the color gradient is from one side to the other. The model from experiment \emph{Seg. \& Proj.}  is used to generate the images.}
  \label{fig:qualitative_results}
  \vspace{-0.5cm}
\end{figure*}

Table \ref{tab:results} displays the results for all experiments. For the TokenPose approach, we evaluate two pretrainings. The pretraining on the COCO dataset with the standard keypoints achieves a better standard PCK than the pretraining with the vectorized keypoint approach from \cite{ludwig2022recognition} on the COCO dataset. This is expected, as TokenPose detects also only the standard keypoints of the skijump dataset. 

Furthermore, we use the \emph{vectorized keypoints} approach with a generation of 5 to 50 arbitrary keypoints for each image and with the improved attention mechanism described in Section \ref{sec:attention}. It achieves good results for the full PCK, the MTE and the PCT, but the PCK on the whole test set decreases by absolute 24.5\% in comparison to TokenPose. 
Therefore, we use the \emph{combined strategies} like described in Section \ref{sec:train_strategy} to improve the standard PCK. In the first experiment (\emph{Std. \& Seg.} in Table \ref{tab:results}), we alternately train on the standard keypoints and the arbitrary points. This leads to nearly the same PCK on the standard keypoints, but the PCT is absolute 1.1\% lower than for the vectorized keypoint approach. Training with the projection keypoints instead of the standard keypoints (experiment \emph{Seg. \& Proj.} in Table \ref{tab:results}) leads to better results. With the vectorized keypoint pretraining, it achieves the same PCT as the vectorized keypoint approach, while the PCK decreases only slightly. 

The results of the \emph{Seg. \& Proj.} seem promising. Therefore, we evaluate this strategy with two pretrainings, the vectorized keypoint pretraining and the standard keypoints pretraining. With the standard keypoints pretraining, the standard PCK is even higher than the standard PCK for the TokenPose approach which is trained only on the standard keypoints. But all other metrics decrease for this experiment. Therefore, we focus on the vectorized keypoints pretraining for all other experiments.

Hence, we use the \emph{Std. \& Proj.} experiment with the vectorized keypoints pretraining in order to generate \emph{pseudo labels}, because it achieves the best results regarding the thickness metrics. We generate 1000 pseudo labels for each image in advance and select 25 of them randomly in each training step. The results of the pseudo label training with all pseudo labels are slightly worse than the results of the other experiments. As we did not use the existing segmentation masks during that experiment, we execute a finetuning on the best weights of this experiment in the vectorized keypoints manner in order to improve the results, but unsuccessful. From the validation curve, we observe a decrease in the standard PCK from step to step. Therefore, we consider a combined training in the next experiment, training alternately on the arbitrary keypoints, the pseudo labels and the standard keypoints (experiment \emph{all PL \& Std. \& Seg.} in Table \ref{tab:results}) or the projection points (experiment \emph{all PL \& Proj. \& Seg.} in Table \ref{tab:results}). Training with the standard keypoints achieves lower scores for all metrics in this case, also for the standard PCK. 

Including pseudo labels in the training process did not lead to better results. A look at the quality of the generated pseudo labels shows that some are wrong. Therefore, we repeat the pseudo label experiments with the best 80\% of the labels. We use a filtering technique based on the standard deviation of the detected keypoints for multiple, differently augmented images like described in Section \ref{sec:pl}. The augmentations that we use are horizontal flipping, 45$^\circ$ rotation (clockwise and counterclockwise) and scaling of 65\% and 135\%. We expected better results with more correct labels, but the results are similar. 

These experiments show that it is most important to have more images to train on. In our case, including pseudo labels does not increase the number of images, because we can use all images already by training on standard keypoints or projection points. Using the projection points results in the best scores because they are more similar to the desired arbitrary keypoints in comparison to the standard keypoints. Figure \ref{fig:qualitative_results} shows some example predictions for different poses.
%-------------------------------------------------------------------------

\section{Conclusion}

This paper proposes a method to detect arbitrary keypoints on the limbs and skis of ski jumpers. We publish a new dataset with annotated images of ski jumpers from 10 TV broadcast videos of ski jumping competitions with a total of 370 jumps in order to provide a public benchmark. We provide annotations for 17 standard keypoints for 2159 images and a test, train and validation split such that each athlete is only contained in one subset. 
%Furthermore, we use detectron2 \cite{detectron2} and DensePose \cite{densepose} to generate segmentation masks. We select 424 usable segmentation masks by hand and include them in the dataset. As the validation set is quite small with only 17 images with segmentation masks, we add some coarsely hand-annotated segmentation masks to enlarge the validation set to 46 images. 
Furthermore, we generate 242 usable segmentation masks and include them in the dataset.
The segmentation masks are only partly correct, many of them contain no or only one ski segmentation mask. Therefore, we cannot finetune a segmentation network in order to generate segmentation masks for all other images. But for our method, this is not a problem, since keypoints are only generated on the available segmentation masks. Problematic are only segmentation masks with wrong borders.

This paper is based on the vectorized keypoint approach presented in \cite{ludwig2022recognition}. For the keypoints on the skis, we modify the technique because the projection points do not necessarily lie in the middle of the skis. Therefore we do not include the line of projection points and only use the intersection points with the segmentation mask. All other keypoints are represented relative to the intersection points. The evaluation metrics for the thickness are adapted accordingly.

Training on the images with available segmentation masks with the vectorized keypoints approach shows two drawbacks. If a lot more keypoint query tokens than during training are used in a single inference step, the detection performance deteriorates. This is an effect of the attention mechanism. In the standard attention, all tokens are correlated with all tokens. Hence, the keypoint query tokens have an influence on each other and on the visual tokens as well. We adapt the attention mechanism in a way that the keypoint query tokens do not have an influence on other keypoint query tokens and on the visual tokens. Only the visual tokens are correlated with each other and with the keypoint query tokens. This solves the problem, as evaluations show. 

The second drawback is the large decrease in the standard PCK, so the detection performance on the standard keypoints is a lot worse. This is caused by the small number of images that the model sees during training. Hence, we experiment with different training strategies on both the segmentation mask dataset and the full dataset. Our experiments show that training jointly on arbitrary and standard keypoints lifts the standard PCK to a large extent, but the PCT and MTE deteriorate. Training on the projection points instead of the standard keypoints leads to better results on these metrics.

Hence, the model proposed in this paper is the first model capable of detecting arbitrary keypoints on the limbs and skis of ski jumpers. Moreover, it can be trained using only a few partly correct segmentation masks.

{\small
\bibliographystyle{ieee_fullname}
\bibliography{egbib}
}

\end{document}